\pdfoutput=1

\documentclass[11pt]{article}

\usepackage{naacl2021}

\usepackage{times}
\usepackage{latexsym}

\usepackage[T1]{fontenc}

\usepackage[utf8]{inputenc}

\usepackage{microtype}

\usepackage{CJKutf8}
\usepackage{multirow}
\usepackage{graphicx}

\title{Frustratingly Easy Edit-based Linguistic Steganography\\ with a Masked Language Model}

\author{Honai Ueoka
  \hspace{13mm}Yugo Murawaki
  \hspace{13mm}Sadao Kurohashi \\
  Graduate School of Informatics, Kyoto University \\
  \texttt{hueoka@icn.cce.i.kyoto-u.ac.jp} \\
  \{\texttt{murawaki}, \texttt{kuro}\}\texttt{@i.kyoto-u.ac.jp} \\
}

\begin{document}
\maketitle
\begin{abstract}
With advances in neural language models, the focus of linguistic steganography has shifted from edit-based approaches to generation-based ones.
While the latter's payload capacity is impressive, generating genuine-looking texts remains challenging.
In this paper, we revisit edit-based linguistic steganography, with the idea that a masked language model offers an off-the-shelf solution.
The proposed method eliminates painstaking rule construction and has a high payload capacity for an edit-based model.
It is also shown to be more secure against automatic detection than a generation-based method while offering better control of the security/payload capacity trade-off.
\end{abstract}

\section{Introduction} \label{sec:intro}

Steganography is the practice of concealing a message in some cover data such that an eavesdropper is not even aware of the existence of the secret message~\citep{simmons1984prisoners,anderson1998limits}.
While images, videos, and audio have been dominant cover media~\citep{fridrich2009steganography}, natural language is a promising choice, thanks to the omnipresence of text~\citep{Bennett2004}.

Formally, the goal of linguistic steganography is to create a steganographic system (\textit{stegosystem}) with which the sender \textit{Alice} encodes a secret message, usually in the form of a bit sequence, into a text and the receiver \textit{Bob} decodes the message, with the requirement that the text is so natural that even if transmitted in a public channel, it does not arouse the suspicion of the eavesdropper \textit{Eve}.
For a stegosystem that creates the text through transformation, we refer to the original text as the \textit{cover text} and the modified text as the \textit{stego text}.
A stegosystem has two objectives, \textit{security} and \textit{payload capacity}.
Security is the degree of how unsuspicious the stego text is while payload capacity is the size of the secret message relative to the size of the stego text.
The two objectives generally exhibit a trade-off relationship~\citep{chang-clark-2014-practical}.

Edit-based approaches used to dominate the research on linguistic steganography.
Arguably, the most effective approach was \textit{synonym substitution}~\citep{Chapman2001,Bolshakov2005,Taskiran2006,chang-clark-2014-practical,Wilson:2016}, where a bit chunk was assigned to each member of a synonym group, for example, `0' to \textit{marry} and `1' to \textit{wed}.
The cover text \textit{She will \underline{marry} him} was then modified to the stego text \textit{She will \underline{wed} him} such that the latter carried the secret bit sequence `1'.

This conceptual simplicity was, however, overshadowed by the complexity of linguistic phenomena such as part-of-speech ambiguity, polysemy, and context sensitivity.
For this reason, edit-based approaches were characterized by the painstaking construction of synonym substitution rules, which were tightly coupled with acceptability checking mechanisms (see \citet{chang-clark-2014-practical} for a review and their own elaborate method).
With all these efforts, edit-based stegosystems suffered from low payload capacity, for example, 2 bits per sentence~\citep{chang-clark-2014-practical}.

\begin{figure*}[t]
    \centering
    \includegraphics[width=0.97\linewidth]{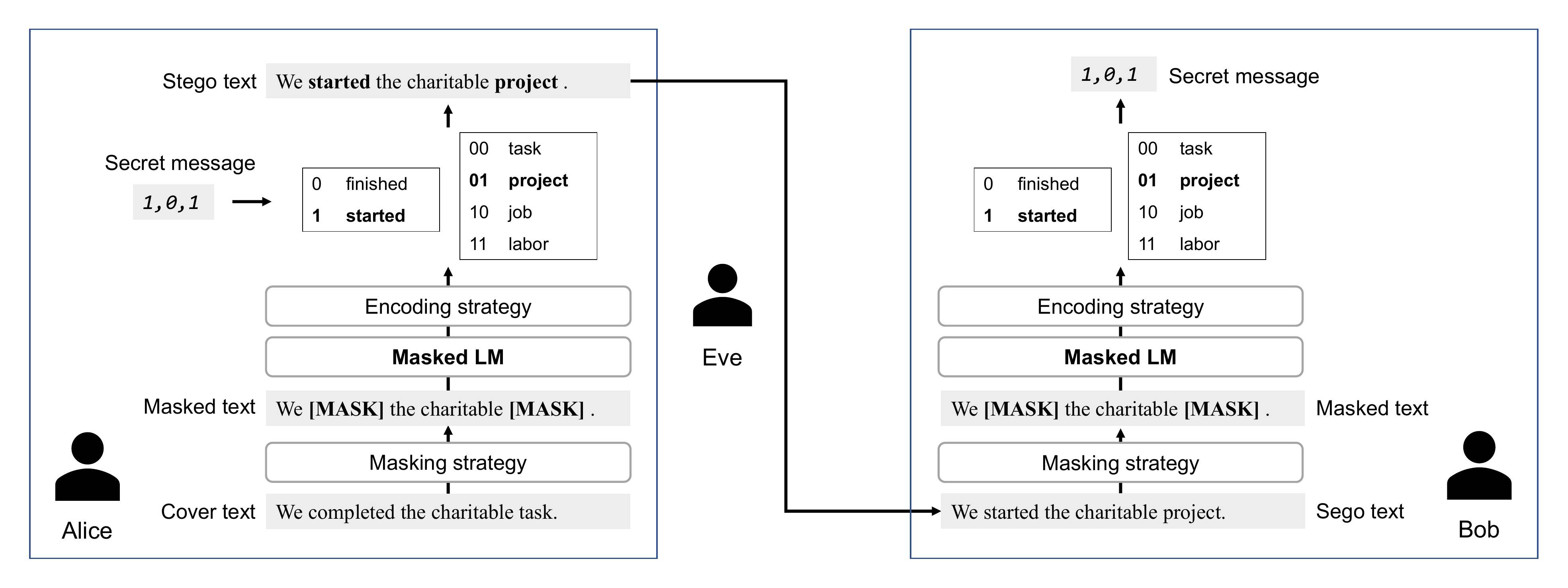}
    \caption{
    Overview of the proposed method.
    Alice (sender) and Bob (receiver) share the masked language model (masked LM) and the masking and encoding strategies in advance.
    Alice masks some tokens in the cover text and makes the masked LM generate a vocabulary distribution for each masked token.
    Bit chunks are assigned to some high-probability subwords in the distribution from which one is chosen according to the secret message.
    The stego text is then transmitted in a public channel Eve (eavesdropper) monitors.
    Receiving the stego text, Bob performs mostly the same procedure to decode the secret message.
    } \label{fig:bert-flow}
\end{figure*}

With advances in neural language models (LMs), edit-based approaches have been replaced by generation-based ones~\citep{fang-etal-2017-generating,Yang:TIFS2019,dai-cai-2019-towards,ziegler-etal-2019-neural,shen-etal-2020-near}.
In these approaches, bit chunks are directly assigned to the conditional probability distribution over the next word estimated by the LM, yielding impressive payload capacities of 1--5 bits per word~\citep{shen-etal-2020-near}.

However, it remains challenging for an LM to generate so genuine-looking texts that they fool both humans and machines~\citep{ippolito-etal-2020-automatic} even if they do not encode secret messages.
It is also worth noting that generation-based stegosystems do not necessarily cut out the need for cover texts, as \citet{ziegler-etal-2019-neural} and \citet{shen-etal-2020-near} conditioned generation on human-written introductory sentences to ensure the stego text quality.

In this paper, we revisit edit-based linguistic steganography.
Our key idea is that a masked language model (masked LM), which was first introduced with BERT~\citep{devlin-etal-2019-bert}, offers an off-the-shelf solution.
Usually treated as an intermediate model with no direct application, the masked LM drastically simplifies an edit-based stegosystem.
It eliminates painstaking rule construction because it readily offers a list of words applicable in the context.
As illustrated in Figure~\ref{fig:bert-flow}, all Alice and Bob have to share is the masking and encoding strategies in addition to the masked LM.

In our experiments, we showed that the proposed method had a high payload capacity for an edit-based model.
As expected, the amount was far smaller than those of generation-based models, but the proposed method offers better control of the security/payload capacity trade-off.
We also demonstrated that it was more secure against automatic detection than a generation-based method although it was rated slightly lower by human adversaries.

Our code is available at \href{https://github.com/ku-nlp/steganography-with-masked-lm}{https://github.com/ku-nlp/steganography-with-masked-lm}.

\section{Proposed Method} \label{sec:proposed}
\subsection{Masked LM} \label{sec:proposed-mlm}
The essential ingredient of the proposed edit-based stegosystem is a masked LM.
It was first introduced along with BERT~\citep{devlin-etal-2019-bert} as an effective pretraining strategy for the Transformer-based~\citep{Vaswani:NIPS2017} neural net.
The pretrained model is usually fine-tuned on downstream tasks, but for our purpose we keep it intact.

Given a text in which some tokens were replaced with the special token \texttt{[MASK]}, the masked LM is trained to recover the original tokens based only on their context.
As a result of the training, it provides a probability distribution over the vocabulary for each masked token according to the applicability in the given context.
Note that high probability items are not necessarily synonymous with the original tokens but nevertheless fit into the context.

Our key insight is that we can use these probability distributions to encode a secret message in the form of a bit sequence.
As shown in Figure~\ref{fig:bert-flow}, Alice and Bob share some \textit{encoding strategy} with which bit chunks are assigned to some high probability items.
Alice creates a stego text by choosing items that correspond to the secret message.
Bob in turn decodes the secret message by selecting bit chunks that correspond to each token in the stego text.
The only remaining requirement for Alice is to share some \textit{masking strategy} with Bob in advance so that Bob can correctly identify the tokens to be masked.

\subsection{Masking Strategy} \label{sec:proposed-masking}

We have various design choices for masking and encoding strategies, which affect both security and payload capacity.
For masked LM training, BERT randomly masked about 15\% of tokens in the input, but we need to ensure that both Alice and Bob mask the same tokens.
In this paper, we present a simple strategy.
As a general rule, we mask every one in $f$ tokens in the input, but we skip tokens if they match any of the following criteria:
\begin{enumerate}
    \item A punctuation or number.
    \item A stopword.
    \item A non-initial subword, which BERT's standard tokenizer marks with the initial ``\#\#''.
\end{enumerate}
Editing subwords is dangerous because there is no 100 percent guarantee that Bob's subword tokenization reproduces Alice's original segmentation.
For example, if ``\#\#break'' in the word ``un \#\#break \#\#able'' is replaced with ``\#us'', the subword tokenizer would segment the new word into ``un \#\#usable'', distorting the masking positions.
We will revisit this problem in Section~ \ref{sec:exp-results}.

The hyperparameter $f$ is expected to control the security/payload capacity trade-off.
A large $f$ lowers the payload capacity but is likely to increase the difficulty of detection.
We also anticipate that since the tokens we decide to skip do not have many good alternatives, not masking them is good for the stego text quality.

\subsection{Encoding Strategy} \label{sec:proposed-encoding}
We use block encoding for simplicity.
For each masked token, we select and sort items whose probabilities are greater than $p$.
To avoid distorting masking positions, we drop items that are to be skipped in the masking phase.
Let $n$ be the largest integer that satisfies $2^n \leq c$, where $c$ is the number of the remaining items.
Each item is given a unique bit chunk of size $n$.
Coding is an active research topic~\citep{dai-cai-2019-towards,ziegler-etal-2019-neural,shen-etal-2020-near} and is orthogonal to our core proposal.

\section{Experiments} \label{sec:exp}
We tested the proposed method with several configurations and compared it with a generation-based method.
To assess security, we employed automatic discriminators and human adversaries.

\subsection{Models and Data} \label{sec:exp-model-data}
\paragraph{BERT}
For the proposed edit-based method, we used BERT~\citep{devlin-etal-2019-bert} as the masked LM.
Specifically, we used Google's \texttt{BERT\textsubscript{Base, Cased}} model and Hugging Face's \texttt{transformers} package~\citep{wolf-etal-2020-transformers} with default settings.
Given a random bit sequence as the secret message and a paragraph as the cover text, the model encoded bit chunks on a sentence-by-sentence basis.
When the bit chunks reached the end of the secret message, the process was terminated, discarding the remaining sentences in the given paragraph.
The last bit chunk usually exceeded the limit, and the remainder was filled with zeros.

\paragraph{GPT-2}
\citet{ziegler-etal-2019-neural} built a state-of-the-art generation-based model on top of the GPT-2 neural LM~\citep{radford2019language}.
We used their original implementation\footnote{\href{https://github.com/harvardnlp/NeuralSteganography}{https://github.com/harvardnlp/NeuralSteganography}} to encode random bit sequences.
We set the option \texttt{finish\_sent} to true to avoid terminating generation at the middle of a sentence.
We tested the temperature parameter $\tau=\{0.4, 0.7, 1.0\}$.
Since the generation was conditioned on context sentences, we supplied the first three sentences of a paragraph.

\paragraph{Data}
We extracted paragraphs from the English part of the CC-100 dataset~\citep{wenzek-etal-2020-ccnet} and used them as the cover texts for BERT and as the contexts for GPT-2.\footnote{
\citet{ziegler-etal-2019-neural} used the CNN/Dailymail~\citep{NIPS2015_afdec700,nallapati-etal-2016-abstractive} as the contexts.

We found, however, that the resulting stego texts were excessively easy for automatic discriminators to distinguish from real news articles, presumably due to domain mismatch with a web corpus on which GPT-2 had been trained.
That is why we chose CC-100, a web corpus, in our experiments.
Note that this setting may have worked slightly against the proposed method because BERT was mainly trained on Wikipedia.
}
For each stegosystem, we also extracted texts that were comparable to the corresponding stego texts in terms of length. We refer to them as \textit{real texts}.

\subsection{Automatic Detection}
We trained discriminators to distinguish stego texts from real texts.
This corresponds to a situation unusually favorable to Eve as she has access to labeled data, though not to secret messages.
A practical reason for this is that after all, we cannot build discriminators without training data.
Besides, a stegosystem's performance is deemed satisfactory if it manages to fool the discriminator even under such disadvantageous conditions.

For each stegosystem, we fine-tuned the same \texttt{BERT\textsubscript{Base, Cased}} model on the binary classification task.
The details are explained in Appendix~\ref{sec:a-auto}.

\begin{table}[t]
    \centering
    \begin{tabular}{c|c|c|c} \hline
        Model & Parameters & Bits/word $\uparrow$  & Acc $\downarrow$  \\ \hline
        BERT &
           \begin{tabular}{c} $f = 3$ \\ $p = 0.02$ \end{tabular}
        & 0.204 & \textbf{0.586} \\ \hline
        GPT-2 & $\tau = 1.0$ & \textbf{1.67} & 0.819 \\ \hline
    \end{tabular}
    \caption{
        Results of automatic detection.
    }
    \label{tab:disc-res}
\end{table}

\begin{table*}[t]
    \centering
    \begin{tabular}{|p{6.7cm}|p{6.7cm}|c|} \hline
        Cover text & Stego text & Rating \\ \hline
        \footnotesize
        Switzerland also has an \textbf{amazing} scientific community that includes \textbf{Geneva} University and CERN, which is one of the top \textbf{research} institutes in the world and is home to the \textbf{world}'s largest particle physics \textbf{laboratory}.
        &
        \footnotesize
        Switzerland also has an \textbf{international} scientific community that includes \textbf{Basel} University and CERN, which is one of the top \textbf{physics} institutes in the world and is home to the \textbf{world}'s largest particle physics \textbf{laboratory}.
        & 5.0 \\ \hline
        \footnotesize
        Allowing local authorities to \textbf{increase} that charge puts the negative \textbf{political} feedback, particularly in areas where \textbf{compliance} is less, like Donegal, on to the local \textbf{councils} and protects the central government.
        &
        \footnotesize
        Allowing local authorities to \textbf{file} that charge puts the negative \textbf{negative} feedback, particularly in areas where \textbf{opposition} is less, like Donegal, on to the local \textbf{government} and protects the central government.
        & 2.8 \\ \hline
    \end{tabular}
    \caption{
    Two examples of stego texts produced by the proposed edit-based method.
    The last column indicates average ratings by crowdworkers.
    }
    \label{tab:bert-example}
\end{table*}

\begin{figure}[t]
    \centering
    \includegraphics[width=0.8\linewidth]{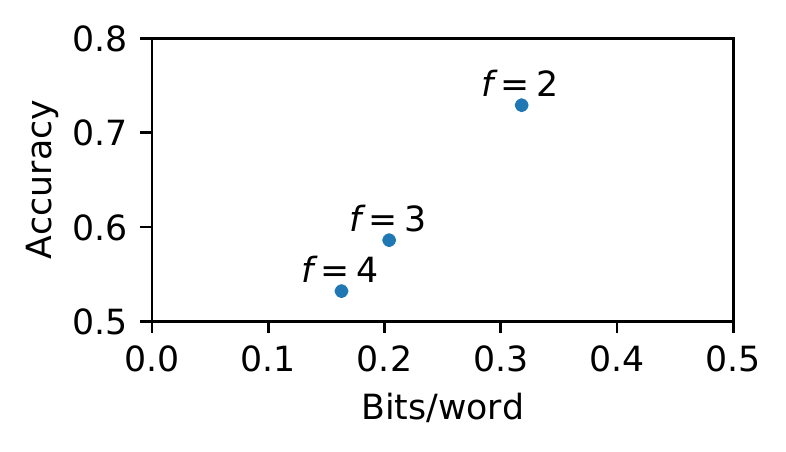}
    \caption{\label{fig:mask-ratio} 
    The effect of the masking interval $f$.
    }
\end{figure}

\begin{figure}[t]
    \centering
    \includegraphics[width=0.8\linewidth]{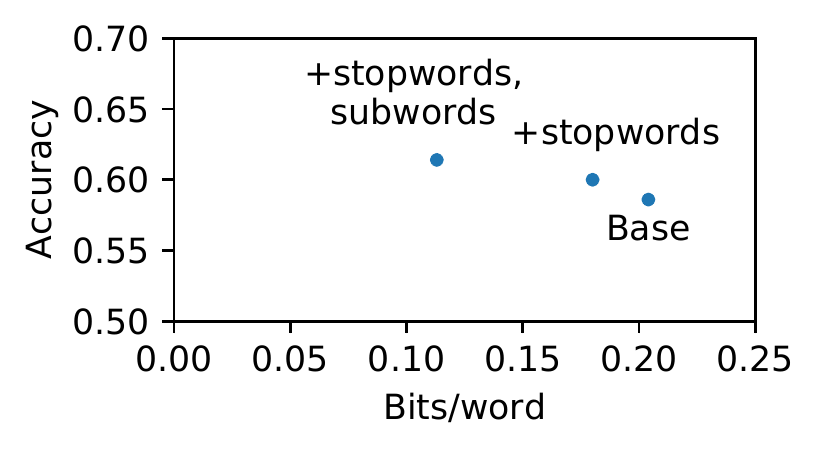}
    \caption{\label{fig:skip-words} 
    The effect of mask skipping heuristics.
    The plus sign indicates that the model stops skipping the specified class of tokens.
    }
\end{figure}

\begin{figure}[t]
    \centering
    \includegraphics[width=0.8\linewidth]{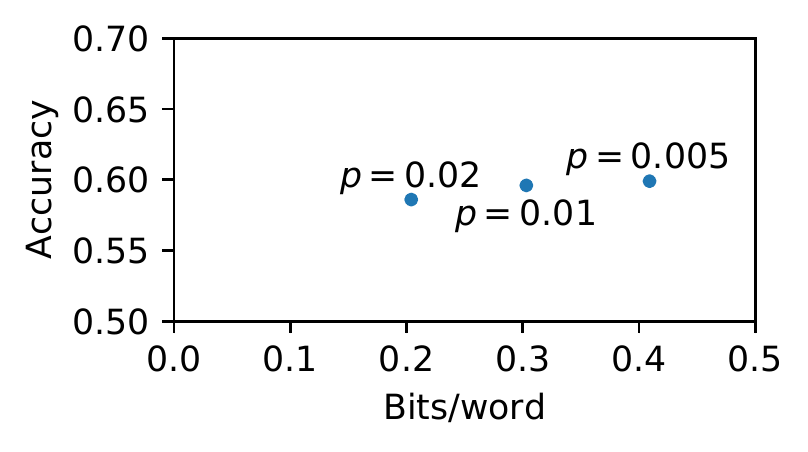}
    \caption{\label{fig:score-threshold} 
    The effect of the probability threshold $p$.
    }
\end{figure}

\begin{table}[t]
    \centering
    \begin{tabular}{|l|l|l|} \hline
        BERT & GPT-2 & Real texts \\
        $4.32\pm 0.97$ & $4.43\pm 0.89$ & $4.54 \pm 0.78$ \\ \hline
    \end{tabular}
    \caption{
    The results of human evaluation.
    The ratings range from 1 to 5, and higher is better. 
    }
    \label{tab:human-score}
\end{table}

\subsection{Human Evaluation}

We asked Amazon Mechanical Turk\footnote{\url{https://www.mturk.com/}} workers to give 5-point scale ratings on the stego and real texts according to naturality and correctness.
Since we found a consistent bias toward shorter texts, we tuned each stegosystem's hyperparameters to generate stego texts with comparable length.
The details are explained in Appendix~\ref{sec:a-human}.

\subsection{Results} \label{sec:exp-results}

Table~\ref{tab:disc-res} shows the result of automatic detection.
As expected, the proposed method, BERT, had a much lower payload capacity than the generation-based GPT-2 although it was high for an edit-based method.
In practical situations, however, security is given priority over payload capacity.
In this respect, BERT's performance was remarkable as its stego texts were nearly indistinguishable from real texts.
By contrast, GTP-2's stego texts were easily detectable for the discriminator even though they were much shorter than BERT's.

Figure~\ref{fig:mask-ratio} shows the effect of the masking interval parameter $f$, with $p = 0.02$.
We can observe a clear trade-off between the two objectives.

Figure~\ref{fig:skip-words} indicates the effectiveness of the mask skipping heuristics explained in Section~\ref{sec:proposed-masking}.
With $f=4$ and $p=0.02$, masking stopwords and subwords not only raised detection accuracy but also lowered payload capacities.
Because these tokens did not have many good alternatives, they consumed only small bit chunks and simply damaged the stego text quality.

As we briefly discussed in Section~\ref{sec:proposed-masking}, editing subwords may cause distortion in mask positions, leading to decoding failures.
We quantified the risk, with the hyperparameter settings of $p = 0.02$ and $f = 3$.
We found that 1.41\% of the masked tokens had substitution candidates that did not reproduce the original segmentations.
Although this danger applies equally to generation-based steganography built on top of subword LMs~\citep{dai-cai-2019-towards,ziegler-etal-2019-neural,shen-etal-2020-near}, to our knowledge, we are the first to point it out.

Figure \ref{fig:score-threshold} shows the effect of the probability threshold $p$.
Lowering the threshold increases the payload capacity because the number of alternative tokens increases.
It did sacrifice detection accuracy, but not as much as we expected.

As for human evaluation, Table~\ref{tab:human-score} summarizes the results with average ratings.
Overall, both methods achieved high average ratings, almost equal to that of the real texts.
However, BERT slightly underperformed GPT-2.
We conjecture that the quality of the cover texts affected the edit-based method more directly than the generation-based method.
Following \citet{ziegler-etal-2019-neural}, we initially used news articles for cover/real texts but switched to web texts because we noticed that the discriminator appeared to exploit the domain mismatch with a web corpus on which GPT-2 had been trained.
Considering the massive quality improvement efforts given to GPT-2's training data, however, there seems to be much room to improve the quality of CC-100~\citep{wenzek-etal-2020-ccnet}.

Table~\ref{tab:bert-example} shows good and bad stego texts produced by the BERT-based method.
In the first example, BERT successfully suggested context-aware words, e.g. \textit{Basel} for a university in Switzerland.
In the second example, a single mistake, the unnatural repetition of \textit{negative}, had a critical impact on human raters.
Finally, we confirmed that the current sentence-wise encoding created a risk of discrepancies between the first and second sentences.

Editing proper nouns like Geneva is prone to factual errors.
One may feel tempted to apply a part-of-speech tagger or a named entity tagger to skip proper nouns.
Just like subword substitution, however, a na\"{i}ve application of automatic analysis does not guarantee the sameness of the masking positions.
A good compromise with a guarantee of success in decoding is to skip words with capitalized letters.
Solving this problem at its source is an interesting direction for future research.

\section{Conclusions} \label{sec:conclusions}

In this paper, we demonstrated that the masked language model could revolutionize edit-based linguistic steganography.
The proposed method is drastically simpler than existing edit-based methods, has a high payload capacity, and allows easy control of the security/payload capacity trade-off.

The masked language model is a general framework adopted by many BERT-like models, of which attempts to handle longer texts~\citep{beltagy2020longformer,wang2020linformer} are particularly relevant to steganography.
Tailoring the training procedure to steganography is also an interesting research direction.

\section*{Ethical Considerations}

This paper works on steganography.
Unlike cryptography, steganography conceals the fact that a secret message is being transmitted as well as its contents.
Steganography can be just fun, but it usually involves a conflict of interest between two parties: those who want to censor media and those who want to evade detection.
Depending on value judgments, either one or both can be evil.
Steganography is an effective tool to counter censorship in countries where encryption is illegal and visibly encrypted messages may be incriminating.
However, it can also be used to transfer malicious data.
As such, steganography can be seen as a dual-use technology.

\bibliography{anthology,custom}
\bibliographystyle{acl_natbib}

\appendix

\section{Details of Automatic Detection} \label{sec:a-auto}

The edit-based and generation-based methods differ considerably in the amount of text needed to encode a secret message because the latter has a higher payload capacity by design.
For a fair comparison, we chose to encode bit sequences of equal size, meaning that the proposed method yielded longer stego texts.
Specifically, we used random 32-bit sequences.
The sequence of sentences was given to the discriminator at once.
For GPT-2, we discarded the contexts and only fed stego texts into the discriminator.

For the binary classification task, we prepared the training, development and test sets with 8,000, 1,000, and 1,000 paragraphs, respectively.
Each set had the same amount of stego and real texts, and they were comparable with respect to length.
The discriminators were trained for 20 epochs, with the batch size of 32.
We applied early stopping if the validation loss did not drop five times in a row.
The model snapshots were saved for every 250 steps, and the one with the lowest validation loss was chosen.
We used Adam (\citet{kingma2017adam}) as the optimizer with the learning late of $10^{-6}$.

\section{Details of Human Evaluation} \label{sec:a-human}

For the proposed edit-based method, we used the same masking and encoding strategies used for automatic detection.
The hyperparameters were as follows: $p = 0.02$, and $f = 4$, stopwords and subwords were skipped.
We used random 6-bit sequences as secret messages.

For the generation-based method, the temperature parameter $\tau$ was set to $0.7$.
We used random 45-bit sequences as secret messages.

We designed a MTurk HIT (human intelligence task) following \citet{abdelnabi2020adversarial}.
Workers were asked to rate texts (each question had 5 texts) with a Likert scale from 1 (lowest) to 5 (highest).
As shown in Table~\ref{tab:human-question}, the ratings were described with the instructions ranging from
\textit{``This sentence is completely understandable, natural, and grammatically correct''} to \textit{``This sentence is completely not understandable, unnatural, and you cannot get its main idea''}.
Each HIT had 5 texts, with stego texts from both methods and real texts of comparable length appearing in a random order.
The questions also had a simple attention check and if the answer to the attention check was wrong, the corresponding HIT was discarded.
We have set the reward per assignment at \$0.3.

\begin{table}[t]
    \centering
    \begin{tabular}{|c|p{5.7cm}|} \hline
        Rating & Description \\ \hline
        5 & The text is understandable, natural, and grammatically and structurally correct. \\
        4 & The text is understandable, but it contains minor mistakes. \\
        3 & The text is generally understandable, but some parts are ambiguous. \\
        2 & The text is mainly not understandable, but you can get the main ideas. \\
        1 & The text is completely not understandable, unnatural, and you cannot get the main ideas. \\ \hline
    \end{tabular}
    \caption{Ratings explanations given in the human evaluation.}
    \label{tab:human-question}
\end{table}

We observed that human raters strongly favored shorter texts.
To verify the observation, we performed linear regression analysis with the number of words as a parameter.
We found that it indeed had a statistically significant negative impact on the ratings with $p < 10^{-3}$ for the t-statistic.

\end{document}